\documentclass[journal]{IEEEtran}

\usepackage{amsmath,amsfonts}

\usepackage{array}
\usepackage{diagbox}
\usepackage[caption=false,font=normalsize,labelfont=sf,textfont=sf]{subfig}
\usepackage[switch]{lineno}
\usepackage{textcomp}
\usepackage{stfloats}
\usepackage{url}
\usepackage{verbatim}
\usepackage{graphicx}
\usepackage{cite}
\PassOptionsToPackage{svgnames}{xcolor}
\usepackage{listings}
\usepackage{graphicx}
\usepackage{overpic} 
\usepackage{multirow}
\usepackage{mathtools}
\usepackage{makecell}
\usepackage{bbding}

\usepackage{amssymb}
\usepackage{microtype}
\usepackage{float}
\usepackage{makecell}
\usepackage{bm}
\usepackage{enumitem}
\usepackage{epstopdf}
\usepackage{booktabs}

\usepackage{algorithmic}
\usepackage[ruled]{algorithm2e}
\usepackage[normalem]{ulem}
\useunder{\uline}{\ul}{}
\usepackage{balance}

\begin{document}

\title{Coarse-to-Fine Video Denoising with Dual-Stage \protect\\ Spatial-Channel Transformer}
\author{Wulian~Yun,~Mengshi~Qi,~\IEEEmembership{Member,~IEEE,}~Chuanming~Wang, ~Huiyuan Fu,~\IEEEmembership{Member,~IEEE,}\\~and~Huadong~Ma,~\IEEEmembership{Fellow,~IEEE}
\thanks{}


}



\maketitle
\begin{abstract}
Video denoising aims to recover high-quality frames from the noisy video. While most existing approaches adopt convolutional neural networks~(CNNs) to separate the noise from the original visual content, however, CNNs focus on local information and ignore the interactions between long-range regions in the frame. Furthermore, most related works directly take the output after basic spatio-temporal denoising as the final result, leading to neglect the fine-grained denoising process. In this paper, we propose a Dual-stage Spatial-Channel Transformer for coarse-to-fine video denoising, which inherits the advantages of both Transformer and CNNs. Specifically, DSCT is proposed based on a progressive dual-stage architecture, namely a coarse-level and a fine-level 
stage to extract dynamic features and static features, respectively. At both stages, a Spatial-Channel Encoding Module is designed to model the long-range contextual dependencies at both spatial and channel levels. Meanwhile, we design a Multi-Scale Residual Structure to preserve multiple aspects of information at different stages, which contains a Temporal Features Aggregation Module to summarize the dynamic representation. Extensive experiments on four publicly available datasets demonstrate our proposed method achieves significant improvements compared to the state-of-the-art methods.
\end{abstract}


\section{Introduction}
\begin{figure}[ht]
    \centering
    \vspace{2.5mm}
    \includegraphics[scale=.55]{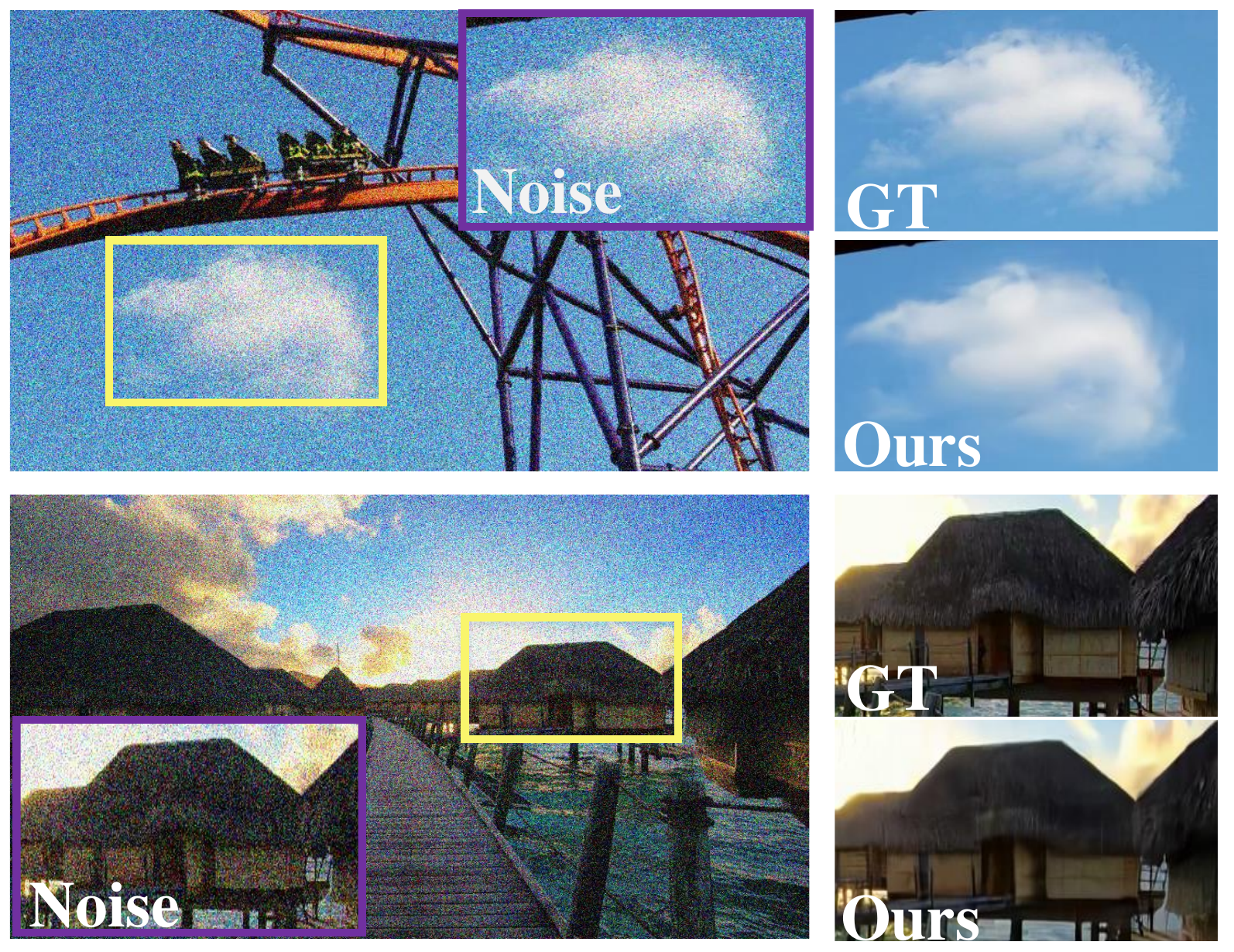}
    \vspace{-0.5mm}
    \caption{Illustration of the video denoising task. From left to right, we show the noisy frames in videos, ground truth clean frames and the predicted results of our proposed method.}
    \label{fig:1}
\end{figure}

\IEEEPARstart{V}{ideo} denoising is a fundamental task in computer vision, which aims to reconstruct clean frames from the noisy video, as shown in Figure~\ref{fig:1}.  
It has attracted considerable research interest because the video denoising can greatly influence the recognition and understanding of video data by subsequent down-stream applications~\cite{9351755,Qi_2020_fewshot,qi2020stc,qi2019sports,qi2019stagnet}, such as vintage film remastering, video surveillance.

Different from the image denoising ~\cite{Zhang2017BeyondAG,Cheng2020NBNetNB,dabov2007image} that focuses on extracting spatial information in the image, the video denoising pays more attention to the spatial information as well as the temporal representation from the video sequence. Therefore, existing methods have been proposed to utilize spatio-temporal information for video denoising. For example, V-BM4D~\cite{Maggioni2012VideoDD} searches similar patches in both the spatial and temporal dimensions, while FastDVDnet~\cite{Tassano2020FastDVDnetTR} utilizes the U-Net architecture~\cite{ronneberger2015u} to process information at both the spatial and temporal level. However, these methods process the spatio-temporal information in a hybrid manner but ignore the spatial noise distribution and temporal deformation. Moreover, these methods only implement the coarse-grained level denoising rather than the fine-grained job.

Nowadays, the CNN-based methods~\cite{claus2019videnn,Tassano2019DVDNETAF,Tassano2020FastDVDnetTR,Davy2018NonLocalVD,Wang2020FirstIT,Chen.2016.DRv} become the mainstream method for video denoising.
However, these CNN-based methods have limitations for capturing long-range relations, because of the intrinsic locality of the convolutional kernel. Alternatively, transformer adopts the self-attention mechanism to capture global contextual relationships and has been shown good performance in natural language processing and computer vision field. Therefore, it is worth exploring how to inherit the advantages of both CNN and transformer for video denoising.

In this study, we propose an end-to-end Dual-stage Spatial-channel Transformer~(DSCT) for coarse-to-fine video denoising. Specifically, DSCT contains a \emph{coarse-level stage} and a \emph{fine-level stage}: the \emph{coarse-level stage} aims to extract the dynamic features from neighbor frames to obtain the coarse-grained denoising result, while the \emph{fine-level stage} transfers the coarse-grained result into the fine-grained clean one by exploiting the static features. During both stages, the Spatial-Channel Encoding Module is introduced to capture the global contextual information from both spatial positions and channels. Moreover, a Multi-Scale Residual Structure is designed to preserve the multi-scale information during both stages, including the temporal information at the \emph{coarse-level stage}, the low-level spatial features and high-level semantic representation at the \emph{fine-level stage}, and intermediate-level context information between the \emph{coarse-level stage} and \emph{fine-level stage}. Additionally, we design a Temporal Features Aggregation Module to summarize the dynamic representation from the video sequence.


The contributions of this paper can be summarized as follows:

\begin{itemize}
\item We propose a novel dual-stage framework for video denoising that gradually obtains the high-quality clean video frames in a coarse-to-fine manner. 

\item We design a Spatial-Channel Transformer to simultaneously capture both the local information and global contextual dependencies between neighbour frames, which contains a Spatial-Channel Encoding Module to extract long-range spatial-channel information.

\item We introduce a Multi-Scale Residual Structure to maintain the multiple aspects of information in different stages and further enhance the representation in each level by adopting skipping residual connections.

\item To verify the effectiveness and superiority of the proposed DSCT, we conduct extensive experiments on four widely-adopted public available datasets, ~\emph{i.e.,} Davis, Set8, Vimeo90k and Raw Videos. The experimental results demonstrate that our model achieves state-of-the-art performance.
\end{itemize}

The rest of this paper is organized as follows. Section~\uppercase\expandafter{\romannumeral2} summarizes recent progresses in Video denoising and Vision transformer. Then, section~\uppercase\expandafter{\romannumeral3} presents the proposed Dual-stage Spatial-channel Transformer for coarse-to-fine video denoising in detail. Afterwards, experimental results and discussions are reported in section~\uppercase\expandafter{\romannumeral4}. Finally, section~\uppercase\expandafter{\romannumeral5} draws the conclusion.

\section{Relate Work}
\subsection{Video Denoising}
Existing video denoising methods can be mainly divided into three categories, including patch-based traditional methods~\cite{Maggioni2012VideoDD,Arias2017VideoDV}, explicit motion compensation methods~\cite{xue2019video,Tassano2019DVDNETAF} and non-explicit motion compensation methods~\cite{Tassano2020FastDVDnetTR,claus2019videnn,Davy2018NonLocalVD}. As an example, V-BM4D~\cite{Maggioni2012VideoDD} is a representative of the patch-based methods, which extends the image denoising method BM3D~\cite{dabov2007image} to the video level. VNLB~\cite{Arias2017VideoDV} is similar to V-BM4D, but its computation time is too longer to be applied in practice. 

With the development of neural networks, deep learning based video denoising methods are proposed in recent years. The first neural network-based method~\cite{Chen.2016.DRv} uses Recurrent Neural Networks (RNNs) for gray-scale image denoising. Xue~\textit{et al.}\cite{xue2019video} propose TOFlow to extract a purpose-built flow representation through aligning frames via CNN. DVDnet~\cite{Tassano2019DVDNETAF} follows independent spatial and temporal steps to model the motion relationship between adjacent frames through optical flow prediction.
However, these explicit motion compensation methods are error-prone and complex, so some methods attempt to perform non-explicit motion compensation in the denoising process. Davy~\textit{et al.}~\cite{Davy2018NonLocalVD} apply CNN to video denoising and use the similarity search strategy for non-local patch processing.
Claus~\textit{et al.}~\cite{claus2019videnn} perform blind noise processing in both the spatial and temporal levels. Considering the complexity of optical flow prediction for DVDnet, Tassano~\textit{et al.}~\cite{Tassano2020FastDVDnetTR} propose FastDVDnet, which exploits U-Net~\cite{ronneberger2015u} blocks. Vaksmanet~\textit{et al.}~\cite{Vaksman_2021_ICCV} introduce patch-craft frames and self-similarity to denoise the video. 

However, above methods are designed to remove synthetic noise such as Gaussian noise, and ignore the realistic noise under real conditions. Therefore, the corresponding work
is now available for realistic noise removal on real datasets. 
Yue~\textit{et al.}~\cite{Yue_2020_CVPR} propose RViDeNet to remove real noise using spatial, channels and temporal correlations, and construct a dynamic video dataset containing real noise. Xu~\textit{et al.}~\cite{9110760} exploit the pixel aggregation at the spatial level, and extend the spatial pixel aggregation at the spatio-temporal level for video denoising. Dewil~\textit{et al}.~\cite{9423377} extend the self-supervised fine-tuning method~\cite{8954460} to a multi-frame denoising network, thus achieving blind video denoising. UDVD~\cite{Sheth_2021_ICCV} combines U-Net and blind-spot~\cite{10.5555/3454287.3454913} network without any explicit motion compensation.

Existing methods mainly use CNNs to extract features, neglecting the long-range context dependencies in video frames. To address the above-mentioned challenges, we propose a new method to inherit the advantages of both transformer and CNNs to capture both global and local information. In addition, we formulate the video denoising task as a coarse-to-fine denoising problem to fully exploit the spatial-temporal information between video frames and further refine the coarse denoising results to fine-grained ones. 

\subsection{Vision Transformer}

Transformer~\cite{vaswani2017attention} is first proposed to solve the machine translation problem in Natural Language Processing (NLP). Since the transformer can capture long-range dependencies in data with self-attention mechanism, it can obtain more global information. Therefore, significant efforts have been made to introduce transformer to computer vision field~\cite{ramachandran2019stand,touvron2021training,carion2020end,cao2021swin,9312201,8578951,10.1007/978-3-030-58452-8_13,9577298,kolesnikov2021image,9710580}. Some methods focus on recognition tasks. For example, Kolesnikov~\textit{et al.}~\cite{kolesnikov2021image} propose ViT, which decomposes an image into a sequence of patches and utilizes transformer for image recognition on the ImageNet dataset. Heo~\textit{et al.}~\cite{Heo2021RethinkingSD} extend the benefits of pooling layers to ViT and propose a Pooling-based Vision Transformer (PiT). Liu~\textit{et al.}~\cite{9710580} propose Swin Transfomer to construct hierarchical structure using sliding window operation. Wang~\textit{et al.}~\cite{9711179} introduce the pyramid structure into the transformer.
Dong~\textit{et al.}~\cite{Dong_2022_CVPR} exploit the cross-shaped window self-attention mechanism to compute the self-attention in horizontal and vertical. 
Other methods focus on restoration tasks. Chen~\textit{et al.}~\cite{Chen_2021_CVPR} propose IPT for various restoration down-stream tasks by jointly training Standard Transformer blocks with multiple tails and heads.
Wang~\textit{et al.}~\cite{Wang2022Uformer} propose a U-Net based architecture with LeWin Transformer block for image restoration tasks such as denoising and deblurring. Zamir~\textit{et al.}~\cite{Zamir_2022_CVPR} propose an efficient Transformer to capture long-range pixel interactions, which can be applied to large images.

Transformer~\cite{Chen_2021_CVPR, Wang2022Uformer,Zamir_2022_CVPR} has been designed for image denoising tasks, but very few are designed for video denoising. In this paper, we propose a Spatial-Channel Transformer, which takes full advantage of the transformer to capture global features in both spatial and temporal dimensions.

\section{Methodology}
\begin{figure*}[htbp]
\begin{center}
\setlength{\fboxrule}{0pt}
\setlength{\fboxsep}{0cm}
\fbox{\rule{0pt}{0in} \rule{.0\linewidth}{0pt}
\hspace{-2.5mm}
     \includegraphics[width=1.0\linewidth]{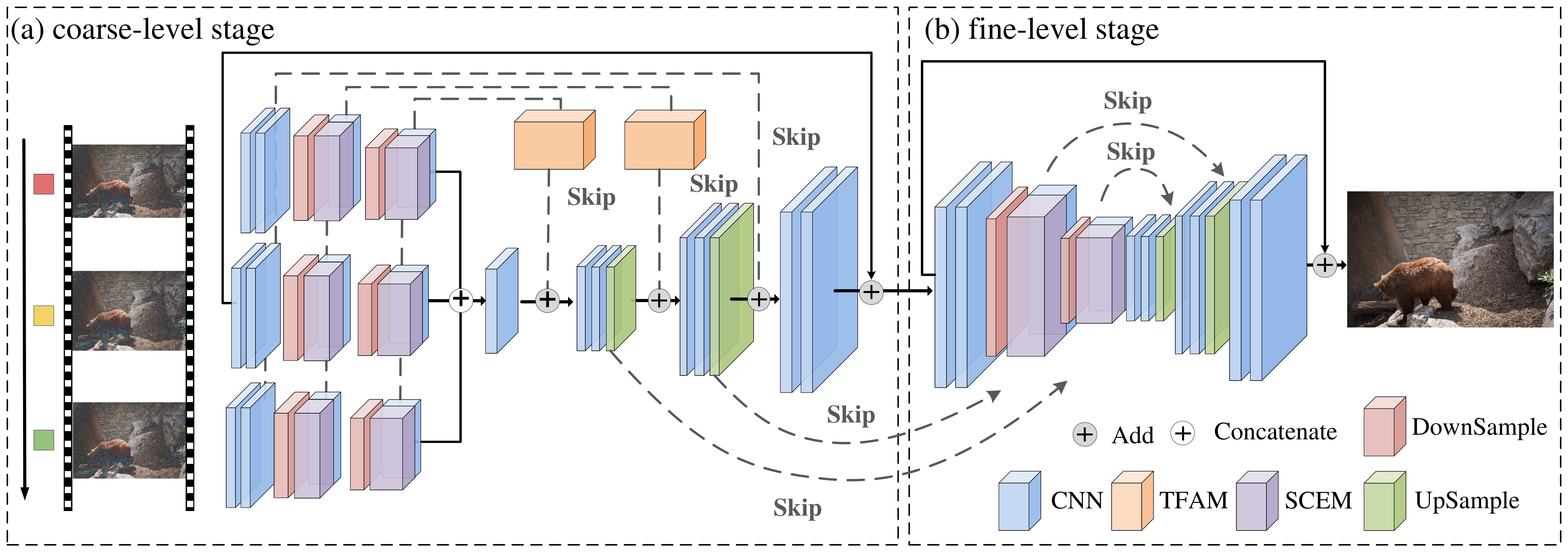}}
\end{center}
\vspace{-0.5mm}
\caption{Overview of our proposed DSCT model. The \emph{coarse-level stage} extracts the dynamic features from the neighbor frames, while the \emph{fine-level stage} extracts the fine-grained static representation from the denoised results. Both stages are constructed in an encoder-decoder architecture. The encoder of each stage contains a Spatial-Channel Encoding Module to extract long-range spatial-channel information. The Multi-Scale Residual Structure is introduced to maintain the multiple aspects of information in different stages, and a Temporal Features Aggregation Module is adopted to summarize the dynamic representation.
}
\label{fig:2}
\end{figure*}

In this work, we propose a Dual-stage Spatial-Channel Transformer for video denoising, as shown in Figure~\ref{fig:2}. DSCT mainly consists of two stages: \emph{coarse-level stage} and \emph{fine-level stage}. The \emph{coarse-level stage} aims at extracting dynamic features from spatio-temporal dimensions of neighbor frames, and then producing the coarse-grained denoising results. While the \emph{fine-level stage} focuses on static features extraction from the coarse-grained denoising results, and generates the boosted fine-grained outputs. In addition, we design a Multi-Scale Residual Structure to maintain the multiple aspects of information in different stages. Note that both stages can be jointly optimized in an end-to-end training architecture. The descriptions of all symbols used in DSCT are shown in Table~\ref{tab:1}.

\begin{table}[h]
\renewcommand{\arraystretch}{1.2}
\setlength{\tabcolsep}{3.1mm}{
\centering
\caption{The list of used symbols and their descriptions in DSCT.}
\vspace{-1.5mm}
\label{tab:1}
\begin{tabular}{c|c}
\specialrule{0.1em}{0.5pt}{0.5pt}
\hline
Symbol & Description \\ \hline
$\tilde{\mathbf{I}}$   &  The noise video frames           \\ \hline
${\mathbf{I}}$     &  The clean video frames        \\ \hline
$\mathbf{I}^{\prime}$       &   The output of the \emph{coarse-level stage}        \\ \hline
$\hat{\mathbf{I}}$    &    The output  of the  \emph{fine-level stage}      \\ \hline
$\mathcal{X}$   &  The features after downsampling 
\\ \hline
$\mathcal{F}^{\text {C}}$    &   The output features of CNN          \\ \hline
${\mathcal{F}}^{\text {SC}}$     &   \makecell{ The output features of  \\ Spatial-Channel Encoding Module and CNN}          \\ \hline
${\mathcal{F}}^{\text {ME}}$     &    The output features of Multi-frame Encoder Module         \\ \hline
$\mathcal{F}^{\text {TF}}$  &   \makecell{ The output features of \\Temporal Features Aggregation
Module}       \\ \hline
$\Omega_{Conv}$   &  The convolution operation of  DSCT model   \\ \hline
$\Omega_{SC}$  & The function of Spatial-Channel Encoding Module  \\ \hline
$\Omega_{SA}$  & Self-Attention  operation\\ \hline
$\Omega_{MSA}$  & Multi-head Self-Attention operation \\ \hline
$\Omega_{D}$ & The function of the decoder at the \emph{coarse-level stage} \\ \hline
 $\Omega_{Fl}$  &  \makecell{The function of encoder and decoder \\ at the  \emph{fine-level stage}}   \\ \hline
$\mathcal{L}$     &       The loss function of DSCT model      \\ \hline
\specialrule{0.1em}{1pt}{1pt}
\end{tabular}
}
\end{table}

\noindent\textbf{Problem Definition:}~Given a training set $\mathcal{I} = \{\tilde{\mathbf{I}}_i, \mathbf{I}_i\}_{i=1}^{N}$, where $\tilde{\mathbf{I}}_i$ denotes the $i$-th noise video frame in $\mathcal{I}$, $\mathbf{I}_i$ is clean frame, and $N$ is the number of $\mathcal{I}$. Our goal is to train a video denoising model based on $\mathcal{I}$, which can transfer each noise frame $\tilde{I}$ into the clean one $I$. 

\subsection{Coarse-Level Stage}

We firstly adopt an encoder-decoder architecture during the \emph{coarse-level stage} to extract the dynamic features from multiple video frames. The \emph{coarse-level stage} is comprised of Multi-frame Encoder Module and Decoder Module, in which the encoder is used to extract the global and local features of each noisy frame by the Spatial-Channel Encoding Module and the CNN. While the Decoder Module is adopted to aggregate the encoded multi-scale features, and transfer them into the coarse-grained denoised results via the CNN and upsample operation.

During this stage, we take the noisy frames set $\{ \tilde{\mathbf{I}}_{i-1}, {\tilde{\mathbf{I}}}_{i},{\tilde{\mathbf{I}}}_{i+1}\}$ as the input, where $\tilde{\mathbf{I}}\in \mathbb{R}^{H \times W \times C}$~(${\mathrm{H}}$, W and C indicate the height, width and input channel number of the frames, respectively). For simplicity, we denote $\tilde{\mathbf{I}}_i$ as the frame that needs to be denoised, while the others $\tilde{\mathbf{I}}_{i-1}$ and $\tilde{\mathbf{I}}_{i+1}$ are denoted as the prior and next video frame, respectively. 

\noindent
\textbf{Multi-frame Encoder Module.}  
Firstly, we use two 3×3 convolution layers followed by BN~\cite{ioffe2015batch} and ReLU~\cite{glorot2011deep} to extract initial features $\mathcal{F}^{\text {C}}$ from different frames:
\begin{equation}
\mathcal{F}^{\text {C}} =\Omega_{Conv}(\tilde{\mathbf{I}}),
\end{equation}
where $\Omega_{Conv}$ means the convolution operation. 

Then, the features passed through two downsample operations, each followed by a Spatial-Channel Encoding Module and two $3\times3$ convolution layers. The features captured from Spatial-Channel Encoding Module and CNN are combined as the output $\mathcal{F}^{\text {SC}}$, which is formulated as:  
\begin{equation}
\mathcal{F}^{\text {SC}}=\Omega_{SC}(\mathcal{X})+\Omega_{Conv}(\mathcal{X}),
\end{equation}
where $\Omega_{SC}$ is the function of Spatial-Channel Encoding Module, $\mathcal{X}$ denotes the feature after downsampling. 

Finally, we obtain the low-level spatial features and high-level dynamic representation of three continuous frames and then fed them into the Decoder Module to obtain the denoising results $\mathbf{I}^{\prime}$ at the \emph{coarse-level stage}.
Among them, the low-level spatial features are the features $\mathcal{F}^{\text {C}}$ of the consecutive input frames after the first pass through CNN. The high-level dynamic representation is the features $\mathcal{F}^{\text {SC}}$ of the consecutive input frames after passing through the Spatial-Channel Encoding Module and CNN. The high-level dynamic representation contains more semantic information than low-level spatial features and helps the network to better understand the content of video frames.

\begin{figure}[htbp]
    \centering
     \includegraphics[scale=.11]{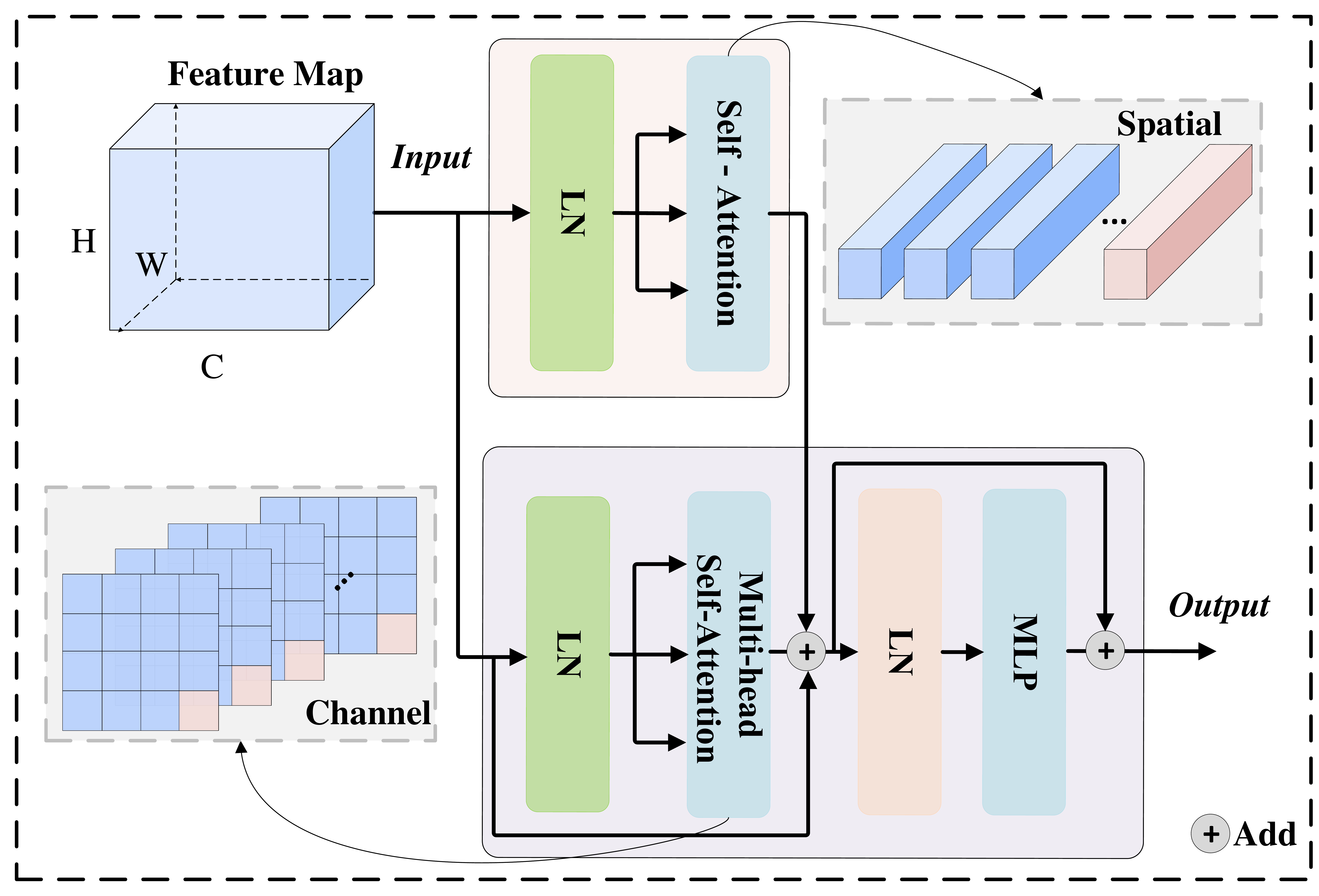}
    \vspace{-0.5mm}
    \caption{Illustration of the proposed Spatial-Channel Encoding Module, which explores global features from both spatial and channel levels.}
    \label{fig:3}
\end{figure}

\noindent
\textbf{Spatial-Channel Encoding Module.} In a given video, different spatial resolution frames contain rich hierarchical representation, and more channels mean more specific information. Therefore, we design a Spatial-Channel Encoding Module~(SCEM) to model the long-range dependencies at both spatial positions and channels. The details of the process is shown in Figure~\ref{fig:3}.

Given a feature $\mathcal{X} \in {\mathbb{R}^{C \times H \times W}}$ as input for the Spatial-Channel Encoding Module, we split it by the patch size of $P \times P$, and then reshape the feature into size $\frac{H W}{P^{2}} \times {P^{2}} \times C$, where $\frac{H W}{P^{2}}$ represents the total number of patches. Firstly, we feed the feature $\mathcal{X} \in \mathbb{R}^{\frac{H W}{P^{2}} \times {P^{2} \times C}}$ through a LayerNorm~(LN)~\cite{Ba2016LayerN} layer, and reshape it to $\mathcal{X} \in \mathbb{R}^{\frac{H W}{P^{2}} \times C \times {P^{2}}}$. Then we compute the Self-Attention (denoted as SA) for feature $\text {X}$, as the following:
\begin{equation}
\begin{aligned}
&\Omega_{\operatorname{SA}}(\cdot)=Softmax(\frac{QK^{T}}{\sqrt{d_{k}}})V,\\
&~~~~~~\hat{\mathcal{X}}=\Omega_{\operatorname{SA}}\left(\mathrm{LN}\left(\mathcal{X}\right)\right),
\end{aligned}
\end{equation}
where $\mathrm{LN}(\cdot)$ denotes the LayerNorm operation, $Q$ is a query matrix, $K$ is a key matrix, $V$ is a value matrix and the size of the attention feature map through softmax is ${C} \times {C}$. 

Meanwhile, Multi-head Self-Attention layer~\cite{vaswani2017attention} (denoted as MSA) takes the feature $\mathcal{X} \in \mathbb{R}^{\frac{H W}{P^{2}} \times {P^{2} \times C}}$ as input, and then calculates the Multi-head Self-Attention as:
\begin{equation}
\begin{aligned}
&\Omega_{\operatorname{MSA}}(\cdot)=\operatorname{Concat}(\Omega_{\operatorname{SA_1}},\cdot\cdot\cdot,\Omega_{\operatorname{SA_h}}){W}^{MSA},\\
&~~~~~~~~~~~~~~\hat{\mathcal{X}^{l}}=\Omega_{\operatorname{MSA}}\left(\mathrm{LN}\left(\mathcal{X}\right)\right),
\end{aligned}
\end{equation}
where $\mathrm{LN}(\cdot)$ denotes the LayerNorm operation,  ${W}^{MSA} \in {\mathbb{R}^{h \cdot {\frac{D}{h}}  \times D}}$ is a fully connected layer, where $h$ denotes the number of head, and the size of the attention feature map through softmax is ${ {P^{2}} \times {P^{2}}}$. In addition, a Multi-Layer Perceptron~(MLP) is also used for further feature transformation followed by one additional LN layer, which can be formulated as follows:
\begin{equation}
\begin{aligned}
&\mathcal{X}=\operatorname{MLP}(\mathrm{LN}({\tilde{\mathcal{X}}}))+{\tilde{\mathcal{X}}}, \\
& ~~~~~where~\tilde{\mathcal{X}}= \hat{\mathcal{X}}+\hat{\mathcal{X}^{l}}+{\mathcal{X}}.
\end{aligned}
\end{equation}

Finally, we reshape the feature back to $\mathcal{X} \in {\mathbb{R}^{C \times H \times W}}$, and obtain the global features containing both spatial and channel information.

\noindent
\textbf{Decoder Module.} 
We pass the output of the Multi-frame Encoder Module through a 1$\times$1 convolution layer, which aims to control the number of the channels as the following:
\begin{equation}
 \mathcal{F}_{M}^{\text {C}} = \Omega_{Conv}(\mathcal{F}_{i-1}^\text{ME}, \mathcal{F}_{i}^\text{ME}, \mathcal{F}_{i+1}^\text{ME}),
\end{equation}
where $\mathcal{F}_{i-1}^\text{ME}$, $\mathcal{F}_{i}^\text{ME}$, $\mathcal{F}_{i+1}^\text{ME}$ denote the output of the Multi-frame Encoding Module. The decoder module exploits the small-scale convolution filters to capture fine-grained cues, and the Pixelshuffle~\cite{7780576} upsampling operation to recover the spatial and channel structure. In the module, we denote two 3$\times$3 convolutional filters and an upsampling operation as an upsampling layer. Following two upsampling layers, two convolution layers are used to smooth and restore the upsampling result as the clean version of the input $\mathbf{I}^{\prime} \in \mathbb{R}^{H \times W \times C}$.
\begin{equation}
\mathbf{I}^{\prime}=\Omega_{D}\left( \mathcal{F}_{M}^{\text {C}}\right),
\end{equation}
where $\Omega_{D}$ denotes the function of the decoder at the \emph{coarse-level stage}.

\subsection{Fine-Level Stage}

Since the proposed \emph{coarse-level stage} aims to extract dynamic features from video neighborhood frames to obtain the coarse-grained results, we further use the \emph{fine-level stage} to extract the static feature from the output of \emph{coarse-level stage} to achieve the fine-grained denoising result.

During the \emph{fine-level stage}, we adopt a similar architecture as the \emph{coarse-level stage}, which includes Spatial-Channel Encoding Module and CNN to extract global and local features. Compared to the \emph{coarse-level stage}, the encoder of \emph{fine-level stage} accepts the output from the decoder of the \emph{coarse-level stage}, so there is only one branch in the encoder and the decoder only uses a simple skip-connection operation here. 

We take the output $\mathbf{I}^{\prime} \in \mathbb{R}^{H \times W \times C}$ of the previous stage as input, and the restored clean video frames $\hat{\mathbf{I}} \in \mathbb{R}^{H \times W \times C}$ can be obtained as the following: 
\begin{equation}
\hat{\mathbf{I}}=\Omega_{Fl}\left(\mathbf{I}^{\prime}\right),
\end{equation}
where $\Omega_{Fl}$ denotes the function of the encoder and decoder at the \emph{fine-level stage}.

\subsection{Multi-Scale Residual Structure}
We design a Multi-Scale Residual Structure to preserve multiple aspects of information during different stages, which contains three different skip operations as follows: 

\noindent(1)~\textbf{Coarse-level stage}: We introduce the skip-connections between decoder and encoder at the \emph{coarse-level stage}, which aims to retain the temporal information from adjacent frames in the Multi-Scale Residual Structure. However, most existing approaches adopt motion compensation or patch-based to employ temporal information, but these methods are error-prone and complex. Then, it is difficult to make full use of video frames by using simple Mean or Conv operations. Mean operation simply stacks features from neighboring frames together without considering the relationship between frames, while Conv operation only focuses on local information between frames, and ignores the correlation between long-range information. 

\begin{figure}[t]
    \centering
    \includegraphics[scale=.0805]{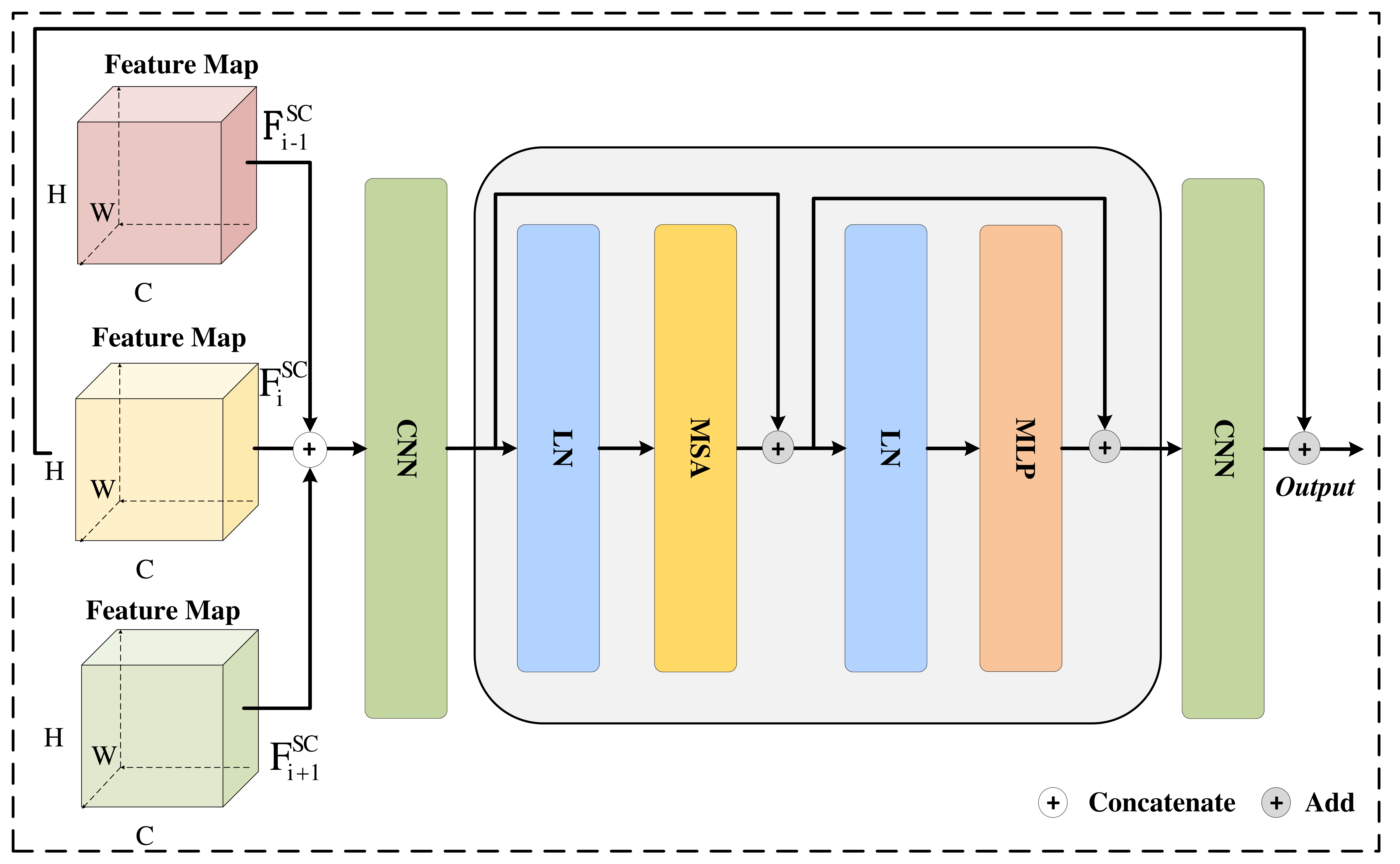}
    \caption{Illustration of the Temporal Features Aggregation Module, which exploits the Self-Attention mechanism to incorporate the global context for different video frames. Among them, the three different color cubes from top to bottom on the left represent the  $\mathcal{F}_ {i-1}^\text{SC}, \mathcal{F}_{i}^\text{SC}, \mathcal{F}_{i+1}^\text{SC}$ features from different frames, respectively.}
    \label{fig:4}
\end{figure}

So, we design the Temporal Features Aggregation Module~(TFAM) to aggregate the temporal features and thus summarize the dynamic representation, by providing additional information for reconstructing target frames and making the restoration process robust, as shown in Figure~\ref{fig:4}. Rather than using explicit motion compensation, we directly perform feature aggregation to implicitly incorporate temporal variation in our method. Among them, we use the Self-Attention operation to model the long-range dependence between frames and fuse the features of adjacent frames together.

In our method, TFAM aggregates the features $\mathcal{F}^\text{SC}$ of adjacent frames on different scales. As shown in Figure 2, TFAM takes the output features $\mathcal{F}_ {i-1}^\text{SC}, \mathcal{F}_{i}^\text{SC}, \mathcal{F}_{i+1}^\text{SC}$ of three consecutive frames through SCEM and CNN as input, where the features are obtained according to Eq.(2). 
The internal processing of TFAM is described as the following:  We first pass the adjacent frames features $\mathcal{F}_ {i-1}^\text{SC}, \mathcal{F}_{i}^\text{SC}, \mathcal{F}_{i+1}^\text{SC}$ through a 3$\times$3 convolution layer, which aims to extract local feature and expand the perception field of input to fit the subsequent operation.
Then, the features $\mathcal{F}_{T}^{\text{C}}$ from the CNN are fed into an MSA layer to obtain the global context information, where the MSA layer is preceded by an LN layer and followed by one MLP layer and one LN layer. 
\begin{equation}
\begin{aligned}
&\tilde{\mathcal{F}_{T}^{\text {C}}}=\operatorname{MSA}(\operatorname{LN}( \mathcal{F}_{T}^{\text {C}}))+ \mathcal{F}_{T}^{\text {C}}, \\
&\hat{\mathcal{F}_{T}^{\text {C}}}=\operatorname{MLP}(\operatorname{LN}( \tilde{\mathcal{F}_{T}^{\text {C}}}))+ \tilde{\mathcal{F}_{T}^{\text {C}}}.
\end{aligned}
\end{equation}

Finally, the size of the channels is controlled by a 3$\times$3 convolutional layer.
And we use a residual operation on the input feature of the intermediate frame and the output of the TFAM. Then, we get the aggregated temporal information $\mathcal{F}^{\text {TF}}$. 

\noindent(2)~\textbf{Between the coarse-level stage and the fine-level stage}:
We introduce the skip-connections to bridge the \emph{coarse-level stage} decoder and the \emph{fine-level stage} encoder. By utilizing such skip-connections, we can introduce intermediate-level context information for the \emph{fine-level stage} encoder to make the network gain more knowledge and thus beneficial to denoise. On the other hand, it can avoid gradient-vanishing due to the deeper network, and ensure the network can back-propagate the gradient properly during the training. 

\noindent(3)~\textbf{Fine-level stage}: We introduced the skip-connections between encoder and decoder at this stage, in order to preserve low-level spatial features and high-level semantic context.

\begin{algorithm}[h]
  \caption{The training process of DSCT.}
  \label{alg:1}
  \LinesNumbered  
  \KwIn{Training dataset $\mathcal{V} = \{\mathcal{I}_j\}_{j=1}^{M}$, training epoch $T$}
  \KwOut{Optimized DSCT model $\mathcal{D}$}
  \For{$t = 1; t \le T ; t++$}
  {
  \For{$j = 1; j \le M ; j++$}
  {
  \For{$i = 1; i \le N ; i++$}
  {
  Generate training set $\{\tilde{\mathbf{I}}_{i-1}, {\tilde{\mathbf{I}}}_{i},{\tilde{\mathbf{I}}}_{i+1}\}$;

  Obtain initial features $\{\mathcal{F}_{i-1}^{\text{C}}, \mathcal{F}_{i}^{\text{C}},\mathcal{F}_ {i+1}^{\text{C}}\}$ by Eq.(1);
 
  Capture global and local features $\{\mathcal{F}_{i-1}^\text{SC}, \mathcal{F}_{i}^{\text{SC}}, \mathcal{F}_{i+1}^{\text{SC}}\}$ by Eq.(2);
 
  Aggregate temporal features $\mathcal{F}^{\text {TF}}$;
 
  Obtain features ~$\{\mathcal{F}_{i-1}^\text{ME}, \mathcal{F}_{i}^\text{ME}, \mathcal{F}_{i+1}^\text{ME}\}$ of Multi-frame Encoder Module;
 
  Generate coarse-grained output $\mathbf{I}^{\prime}_{i}$ by Eq.(7);

  Generate final denoising video  $\hat{\mathbf{I}_{i}}$ by Eq.(8);

  Optimize $\mathcal{D}$ minimizing  Eq.(10);
   }
  }
 }
\end{algorithm}

\subsection{Loss Function}
Finally, the proposed whole network can be trained~(the training process is described in Alg.~\ref{alg:1}) in an end-to-end manner with the $L_{2}$ loss, formulated as the following:
\begin{equation} 
\mathcal{L}=\frac{1}{2 N} \sum_{i=1}^{N}\left\|\hat{\mathbf{I}}_{i}-{\mathbf{I}}_{i}\right\|^{2} ,
\end{equation}
where $\hat{\mathbf{I}}_{i}$ and ${\mathbf{I}}_{i}$ indicate the denoising frame result and ground truth clean frame, respectively, and $N$ is the number of the training data.

\section{Experiments}

To verify the effectiveness of our proposed method, we conduct extensive experiments on four widely-adopted datasets, ~\emph{i.e.}, Davis~\cite{khoreva2018video}, Set8~\cite{Tassano2020FastDVDnetTR}, Vimeo90k~\cite{xue2019video} and Raw Videos~\cite{Yue_2020_CVPR}. Note that we train our model on Davis, then test our model on Davis-test, Set8 and Vimeo90k-test. To evaluate the generalization performance, we test our model on the real dynamic Raw videos dataset~\cite{Yue_2020_CVPR}.

\subsection{Datasets and Settings}

\noindent
\textbf{Davis~\cite{khoreva2018video}} is a high quality and resolution video segmentation dataset. It contains both 480p and 1080p resolution densely annotated videos, and we chose 480p resolution data for training and testing. In practice, we select 90 video sequences as the training set and another 30 sequences for test. 

\noindent
\textbf{Set8~\cite{Tassano2020FastDVDnetTR}} includes 8 video sequences of size 960$\times$540, of which four sequences are collected from Derf’s Test Media collection and another four from the GoPro camera. 

\noindent
\textbf{Vimeo90k~\cite{xue2019video}} is a video dataset covering a variety of real-world scenes and actions, which consists of 89,000 videos of size 256$\times$448. Vimeo90k-test is used for evaluation including 7824 sequences and each sequence contains 7 frames.

\noindent
\textbf{Raw videos~\cite{Yue_2020_CVPR}} is a real noisy video dataset that has a total of 11 different videos of indoor scenes, and each video consists of 7 frames. The different scenes are captured using cameras with five different ISO levels from 1600 to 25600, leading to different levels of noise in videos.

\noindent
\textbf{Metrics.} We use two quantitative measures to evaluate the performance: Peak Signal-to-Noise Ratio~(PSNR) and Spatio-Temporal Reduced Reference Entropic Differences~(ST-RRED)~\cite{Soundararajan2013VideoQA}.

\noindent
\textbf{Compared Methods.}~To evaluate the effectiveness of the proposed DSCT, we compare our model with the state-of-the-art video denoising and image denoising methods. The video denoising methods include V-BM4D~\cite{Maggioni2012VideoDD}, TOFlow~\cite{xue2019video}, VNLB~\cite{Arias2017VideoDV}, VNLnet~\cite{Davy2018NonLocalVD}, FastDVDnet~\cite{Tassano2020FastDVDnetTR}, DVDnet~\cite{Tassano2019DVDNETAF}, UDVD\cite{Sheth_2021_ICCV}, RViDeNet~\cite{Yue_2020_CVPR}, and PaCNet~\cite{Vaksman_2021_ICCV}. Image denoising methods include DIDN~\cite{9025411}, DnCNN~\cite{Zhang2017BeyondAG} and NBNet~\cite{Cheng2020NBNetNB}, which take each frame of the video as input to the model during training and evaluation.

\subsection{Implementation Details}  
We implement our DSCT with PyTorch \cite{paszke2019pytorch} and GeForce RTX 2080 Ti GPU. We train our model with Adam optimizer\cite{2014Adam} with $\beta_{1}$ = 0.9, $\beta_{2}$ = 0.999 and 100 epochs, and the initial value of the learning rate is set as 0.001, with the learning rate decays by a factor of 10 at the 50,60,80 epoch in turn. The batch size is set to 64. The training frames are cropped into 96$\times$96 RGB patches and the contiguous frames are cropped at the same location. During training, we augment the training data by random horizontal flips and vertical flips with random rotation scale factors of $90^{\circ}$, $180^{\circ}$ and $270^{\circ}$. The patch size and head size are set to 4$\times$4 and 4. Following the same setting in~\cite{Tassano2020FastDVDnetTR}, each sequence is limited to a maximum of 85 frames during the evaluation stage.

\subsection{Evaluation}

\noindent
\textbf{Results on Gaussian noise.}~We evaluate the denoising performance of our proposed model on different nature video datasets. During training, we train the model with Davis training set by adding additive white Gaussian noise~(AWGN) of $\sigma \in[5,50]$ to the clean video sequence. During testing, we add Gaussian noise with five representative standard deviations $\sigma$ = 10, 20, 30, 40, and 50 for the test set. Table~\ref{tab:2} shows the evaluation results in terms of PSNR and ST-RRED on Davis, Set8 and Vimeo90k datasets. We can find our proposed DSCT outperforms the state-of-the-art video/image denoising methods under different noise w.r.t PSNR, and achieves competitive results w.r.t ST-RRED on Davis and Set8 dataset. We explain the improved performance as the following reasons: firstly, the \emph{coarse-level stage} can fuse the effective information of the noised frame with its neighbors, and the \emph{fine-level stage} can fully exploit the hidden information in the coarse-level denoising results to generate the more clean version than the coarse-level one; secondly, our method inherits the advantage of transformer and CNNs, which can effectively model the long-range dependency and local perception at the same time. 

Furthermore, we conducted experiments on Vimeo90k dataset to verify the generality of our approach. We show the evaluation results on Table \ref{tab:2} by comparing our proposed method with DnCNN~\cite{Zhang2017BeyondAG}, V-BM4D \cite{Maggioni2012VideoDD}, TOFLow \cite{xue2019video}, and FastDVDnet~\cite{Tassano2020FastDVDnetTR}. We can observe that our method can obtain the best results in terms of PSNR. Meanwhile, with the increasing of noise standard, our model can still obtain satisfied results. For example, the PSNR of our proposed method can reach to 33.46 at the noise standard of 50, demonstrating our method has better generality than other competitors. 
However, FastDVDnet~\cite{Tassano2020FastDVDnetTR} achieves slightly better results w.r.t ST-RRED than our method, because they utilize more reference frames than we used to recover the current frame. 

\begin{center}

\begin{table*}[ht]
\renewcommand{\arraystretch}{1.4}
\setlength{\tabcolsep}{3.7mm}{
\centering
\caption{PSNR/ST-RRED comparison of the state-of-the-art denoising methods on Davis, Set8 and Vimeo90k datasets. The best performance is highlighted in bold.}

\vspace{-1.4mm}
\label{tab:2}
\begin{tabular}{|c|c|cl|cl|cl|cl|cl|}
\hline
\multirow{2}{*}{Dataset}   & \multirow{2}{*}{Method} & \multicolumn{2}{c|}{$\sigma=10$}          & \multicolumn{2}{c|}{$\sigma=20$}          & \multicolumn{2}{c|}{$\sigma=30$}          & \multicolumn{2}{c|}{$\sigma=40$}          & \multicolumn{2}{c|}{$\sigma=50$}           \\ \cline{3-12} 
                           &                        & \multicolumn{2}{c|}{PSNR/ST-RRED} & \multicolumn{2}{c|}{PSNR/ST-RRED} & \multicolumn{2}{c|}{PSNR/ST-RRED} & \multicolumn{2}{c|}{PSNR/ST-RRED} & \multicolumn{2}{c|}{PSNR/ST-RRED}  \\ \hline
\multirow{10}{*}{Davis}    & DnCNN~\cite{Zhang2017BeyondAG}                  & \multicolumn{2}{c|}{38.28/3.26}  & \multicolumn{2}{c|}{34.74/11.75} & \multicolumn{2}{c|}{32.74/24.85} & \multicolumn{2}{c|}{31.34/42.04} & \multicolumn{2}{c|}{30.27/63.50}  \\
                          & NBNet~\cite{Cheng2020NBNetNB}                 & \multicolumn{2}{c|}{32.83/6.86}  & \multicolumn{2}{c|}{29.94/23.44} & \multicolumn{2}{c|}{28.22/48.20} & \multicolumn{2}{c|}{26.91/80.13} & \multicolumn{2}{c|}{25.80/119.99} \\
                          & V-BM4D~\cite{Maggioni2012VideoDD}                 & \multicolumn{2}{c|}{37.58/4.26}  & \multicolumn{2}{c|}{33.88/11.02} & \multicolumn{2}{c|}{31.65/21.91} & \multicolumn{2}{c|}{30.05/36.60} & \multicolumn{2}{c|}{28.80/54.82}  \\
                          & VNLB~\cite{Arias2017VideoDV}                  & \multicolumn{2}{c|}{38.85/3.22}  & \multicolumn{2}{c|}{35.68/6.77}  & \multicolumn{2}{c|}{33.73/\textbf{12.08}} & \multicolumn{2}{c|}{32.32/19.33} & \multicolumn{2}{c|}{31.13/28.21}  \\
                          & VNLnet~\cite{Davy2018NonLocalVD}                & \multicolumn{2}{c|}{35.83/2.81}  & \multicolumn{2}{c|}{34.49/\textbf{6.11}}  & \multicolumn{2}{c|}{ - }           & \multicolumn{2}{c|}{32.32/18.63} & \multicolumn{2}{c|}{31.43/28.67}  \\
                          & DVDnet~\cite{Tassano2019DVDNETAF}                 & \multicolumn{2}{c|}{38.13/4.28}  & \multicolumn{2}{c|}{35.70/7.54}  & \multicolumn{2}{c|}{34.08/12.19} & \multicolumn{2}{c|}{32.86/\textbf{18.16}} & \multicolumn{2}{c|}{31.85/\textbf{25.63}} \\
                          & FastDVDnet~\cite{Tassano2020FastDVDnetTR}             & \multicolumn{2}{c|}{38.71/3.49}  & \multicolumn{2}{c|}{35.77/7.46}  & \multicolumn{2}{c|}{34.04/13.08} & \multicolumn{2}{c|}{32.82/20.93} & \multicolumn{2}{c|}{31.86/28.89}  \\
                          & PaCNet~\cite{Vaksman_2021_ICCV}                 & \multicolumn{2}{c|}{39.97/\;\;- }     & \multicolumn{2}{c|}{36.82/\;\;- }     & \multicolumn{2}{c|}{34.79/\;\;- }     & \multicolumn{2}{c|}{33.34/\;\;- }     & \multicolumn{2}{c|}{32.20/\;\;- }      \\
                          & UDVD~\cite{Sheth_2021_ICCV}                   & \multicolumn{2}{c|}{-}           & \multicolumn{2}{c|}{ - }           & \multicolumn{2}{c|}{33.92/\;\;- }     & \multicolumn{2}{c|}{32.68/\;\;- }     & \multicolumn{2}{c|}{31.70/\;\;- }      \\
                           \cline{2-12} 
                          & \textbf{DSCT~(Ours)}                   & \multicolumn{2}{c|}{\textbf{40.19}/\textbf{2.30}}  & \multicolumn{2}{c|}{\textbf{36.96}/6.77}  & \multicolumn{2}{c|}{\textbf{35.08}/13.09} & \multicolumn{2}{c|}{\textbf{33.74}/21.23} & \multicolumn{2}{c|}{\textbf{32.69}/31.06}  \\ \hline
\multirow{10}{*}{Set8}     & DnCNN~\cite{Zhang2017BeyondAG}                  & \multicolumn{2}{c|}{36.22/2.72}  & \multicolumn{2}{c|}{32.69/10.23} & \multicolumn{2}{c|}{30.72/22.70} & \multicolumn{2}{c|}{29.36/39.74} & \multicolumn{2}{c|}{28.32/60.41}  \\
                          & NBNet~\cite{Cheng2020NBNetNB}                  & \multicolumn{2}{c|}{31.15/6.38}  & \multicolumn{2}{c|}{28.17/22.40} & \multicolumn{2}{c|}{26.47/45.87} & \multicolumn{2}{c|}{25.24/75.06} & \multicolumn{2}{c|}{24.26/108.53} \\
                          & V-BM4D~\cite{Maggioni2012VideoDD}                 & \multicolumn{2}{c|}{36.05/3.87}  & \multicolumn{2}{c|}{32.19/9.89}  & \multicolumn{2}{c|}{30.00/19.58} & \multicolumn{2}{c|}{28.48/32.82} & \multicolumn{2}{c|}{27.33/49.20}  \\
                          & VNLB~\cite{Arias2017VideoDV}                  & \multicolumn{2}{c|}{\textbf{37.26}/2.86}  & \multicolumn{2}{c|}{33.72/6.28}  & \multicolumn{2}{c|}{31.74/\textbf{11.53}} &\multicolumn{2}{c|}{30.39/18.57} & \multicolumn{2}{c|}{29.24/27.39}  \\
                          & VNLnet~\cite{Davy2018NonLocalVD}                 & \multicolumn{2}{c|}{37.10/3.43}  & \multicolumn{2}{c|}{33.88/6.88}  & \multicolumn{2}{c|}{ - }           & \multicolumn{2}{c|}{30.55/19.71} & \multicolumn{2}{c|}{29.47/29.78}  \\
                          & DVDnet~\cite{Tassano2019DVDNETAF}                & \multicolumn{2}{c|}{36.08/4.16}  & \multicolumn{2}{c|}{33.49/7.54}  & \multicolumn{2}{c|}{31.79/12.61} & \multicolumn{2}{c|}{30.55/19.05} & \multicolumn{2}{c|}{29.56/27.97}  \\
                         & FastDVDnet~\cite{Tassano2020FastDVDnetTR}             & \multicolumn{2}{c|}{36.44/3.00}  & \multicolumn{2}{c|}{33.43/6.65}  & \multicolumn{2}{c|}{36.18/11.85} & \multicolumn{2}{c|}{30.64/\textbf{18.45}} & \multicolumn{2}{c|}{29.53/\textbf{26.75}}  \\
                          & PaCNet~\cite{Vaksman_2021_ICCV}                 & \multicolumn{2}{c|}{37.06/\;\;- }     & \multicolumn{2}{c|}{33.94/\;\;- }     & \multicolumn{2}{c|}{32.05/\;\;- }     & \multicolumn{2}{c|}{30.70/\;\;- }     & \multicolumn{2}{c|}{29.66/\;\;- }      \\
                          & UDVD~\cite{Sheth_2021_ICCV}                   & \multicolumn{2}{c|}{ - }           & \multicolumn{2}{c|}{ - }           & \multicolumn{2}{c|} {32.01/\;\;- }     & \multicolumn{2}{c|}{\textbf{30.92}/\;\;- }     & \multicolumn{2}{c|}{29.89/\;\;- }      \\
 \cline{2-12} 
                          & \textbf{DSCT~(Ours)}                    & \multicolumn{2}{c|}{37.02/ \textbf{2.06} }  & \multicolumn{2}{c|}{\textbf{33.96}/\textbf{6.26}}  & \multicolumn{2}{c|}{\textbf{32.16}/12.12} & \multicolumn{2}{c|}{30.91/19.68} & \multicolumn{2}{c|}{\textbf{29.93}/28.83}  \\ \hline
\multirow{5}{*}{Vimeo90k} & DnCNN~\cite{Zhang2017BeyondAG}                   & \multicolumn{2}{c|}{39.41/4.83}  & \multicolumn{2}{c|}{36.08/17.41} & \multicolumn{2}{c|}{34.06/36.72} & \multicolumn{2}{c|}{32.60/62.10} & \multicolumn{2}{c|}{31.45/92.63}  \\
                          & V-BM4D~\cite{Maggioni2012VideoDD}                 & \multicolumn{2}{c|}{40.23/3.93}  & \multicolumn{2}{c|}{36.22/11.60} & \multicolumn{2}{c|}{33.71/22.87} & \multicolumn{2}{c|}{31.86/37.46} & \multicolumn{2}{c|}{30.40/55.17}  \\
                          & TOFlow~\cite{xue2019video}                & \multicolumn{2}{c|}{31.45/30.62} & \multicolumn{2}{c|}{34.43/27.43} & \multicolumn{2}{c|}{31.76/28.33} & \multicolumn{2}{c|}{27.87/32.84} & \multicolumn{2}{c|}{24.92/41.32}  \\
                          & FastDVDNet~\cite{Tassano2020FastDVDnetTR}             & \multicolumn{2}{c|}{38.87/\textbf{2.93}}  & \multicolumn{2}{c|}{36.15/\textbf{8.13}}  & \multicolumn{2}{c|}{34.40/\textbf{15.15}} & \multicolumn{2}{c|}{33.09/\textbf{24.02}} & \multicolumn{2}{c|}{32.05/\textbf{34.62}}  \\ \cline{2-12} 
                          & \textbf{DSCT~(Ours)}                    & \multicolumn{2}{c|}{\textbf{40.77}/3.11}  & \multicolumn{2}{c|}{\textbf{37.81}/8.65}  & \multicolumn{2}{c|}{\textbf{35.95}/16.32} & \multicolumn{2}{c|}{\textbf{34.56}/25.99} & \multicolumn{2}{c|}{\textbf{33.46}/37.73}  \\ \hline

\end{tabular}
}
\end{table*}
\end{center}

\begin{figure*}[t]
\begin{center}
\setlength{\fboxrule}{0pt}
\setlength{\fboxsep}{0cm}
\fbox{\rule{0pt}{0in} \rule{.0\linewidth}{0pt}
\hspace{-2.5mm}
     \includegraphics[width=0.99\linewidth]{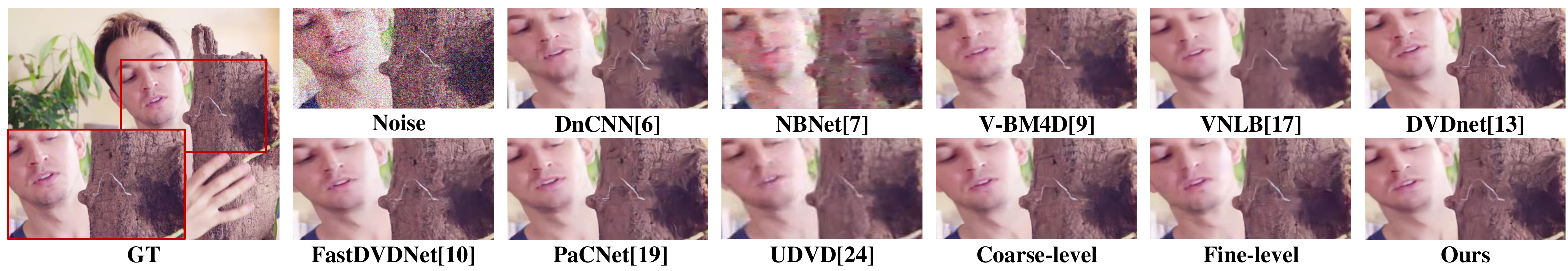}}
\end{center}
\vspace{-2.5mm}
\caption{Denoising visualized results with Gaussian noise level $\sigma$ = 50 on Davis dataset of the proposed DSCT, coarse-level stage, fine-level stage and other state-of-the-art methods.
}
\label{fig:5}
\end{figure*}

\begin{figure*}[t]
\begin{center}
\setlength{\fboxrule}{0pt}
\setlength{\fboxsep}{0cm}
\fbox{\rule{0pt}{0in} \rule{.0\linewidth}{0pt}
\hspace{-2.5mm}
     \includegraphics[width=0.98\linewidth]{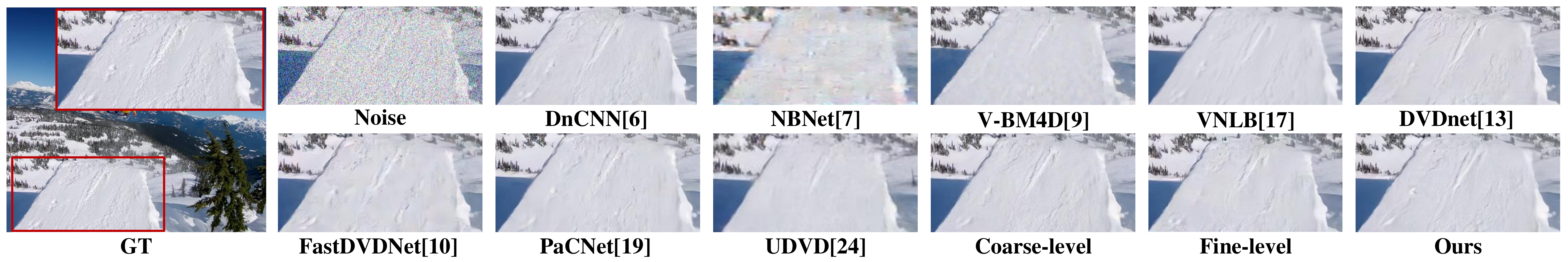}}
\end{center}
\vspace{-3.5mm}
\caption{Denoising visualized results with Gaussian noise level $\sigma$ = 50 on Set8 dataset of the proposed DSCT, coarse-level stage, fine-level stage and other state-of-the-art methods.
}
\label{fig:6}
\end{figure*}

\noindent
\textbf{Visualization results on Davis and Set8.}~ 
Regarding the qualitative results, we show a few of visual results on Davis and Set8 datasets at $\sigma=50$ in Figure~\ref{fig:5} and Figure~\ref{fig:6}, respectively. 
We observe that our method recovers more fine-grained and clearer details in the frame than other methods, especially the visual appearance of the bud part of the tree trunk, the nose of the face, and the bumpy of the snow traces. While the videos restored by another methods have more blurry background. The better visualization results indicate that we combine local and global information from video frames for denoising can generate natural and smooth visual content. Besides, it demonstrates that our proposed Multi-scale Residual Structure can reconstruct enough details. Furthermore, the visual results of the dual-stage are clearer than the results only with the \emph{coarse-level stage} or the \emph{fine-level stage}, which intuitively indicates that our proposed coarse-to-fine strategy is very useful.

\begin{figure*}[ht]
\begin{center}
\setlength{\fboxrule}{0pt}
\setlength{\fboxsep}{0cm}
\fbox{\rule{0pt}{0in} \rule{.0\linewidth}{0pt}
\hspace{-3.5mm}
     \includegraphics[width=0.98\linewidth]{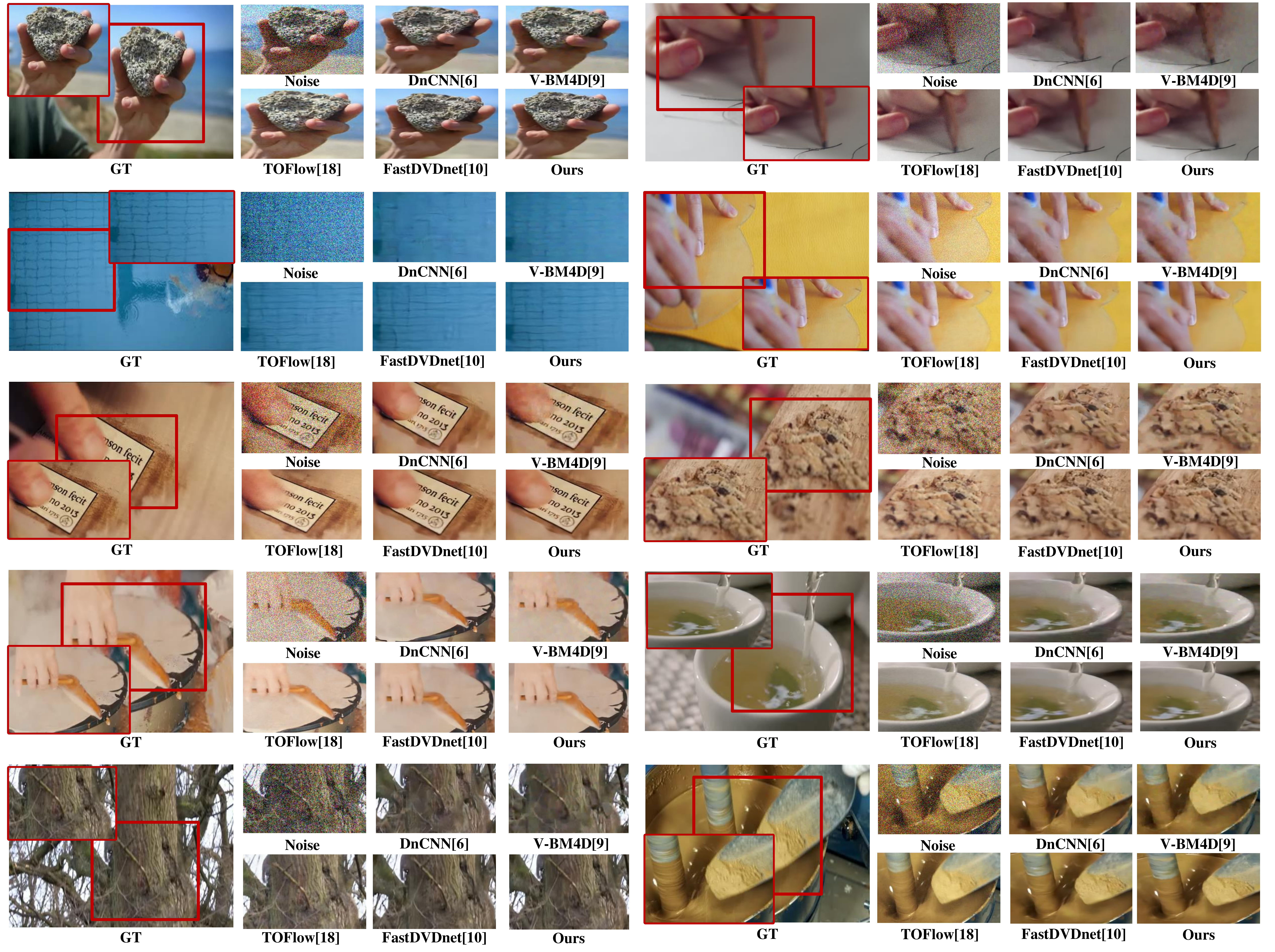}}
\end{center}
\vspace{-4.5mm}
\caption{Denoising visualized results with Gaussian noise level $\sigma$ = 50 on Vimeo90k dataset of the proposed DSCT and other state-of-the-art methods. From top to bottom shows different frames in a given video. }
\label{fig:7}
\end{figure*}  
\begin{figure}[ht]
\begin{center}
\setlength{\fboxrule}{0pt}
\setlength{\fboxsep}{0cm}
\fbox{\rule{0pt}{0in} \rule{.0\linewidth}{0pt}
\hspace{-3.5mm}
     \includegraphics[width=1.0\linewidth]{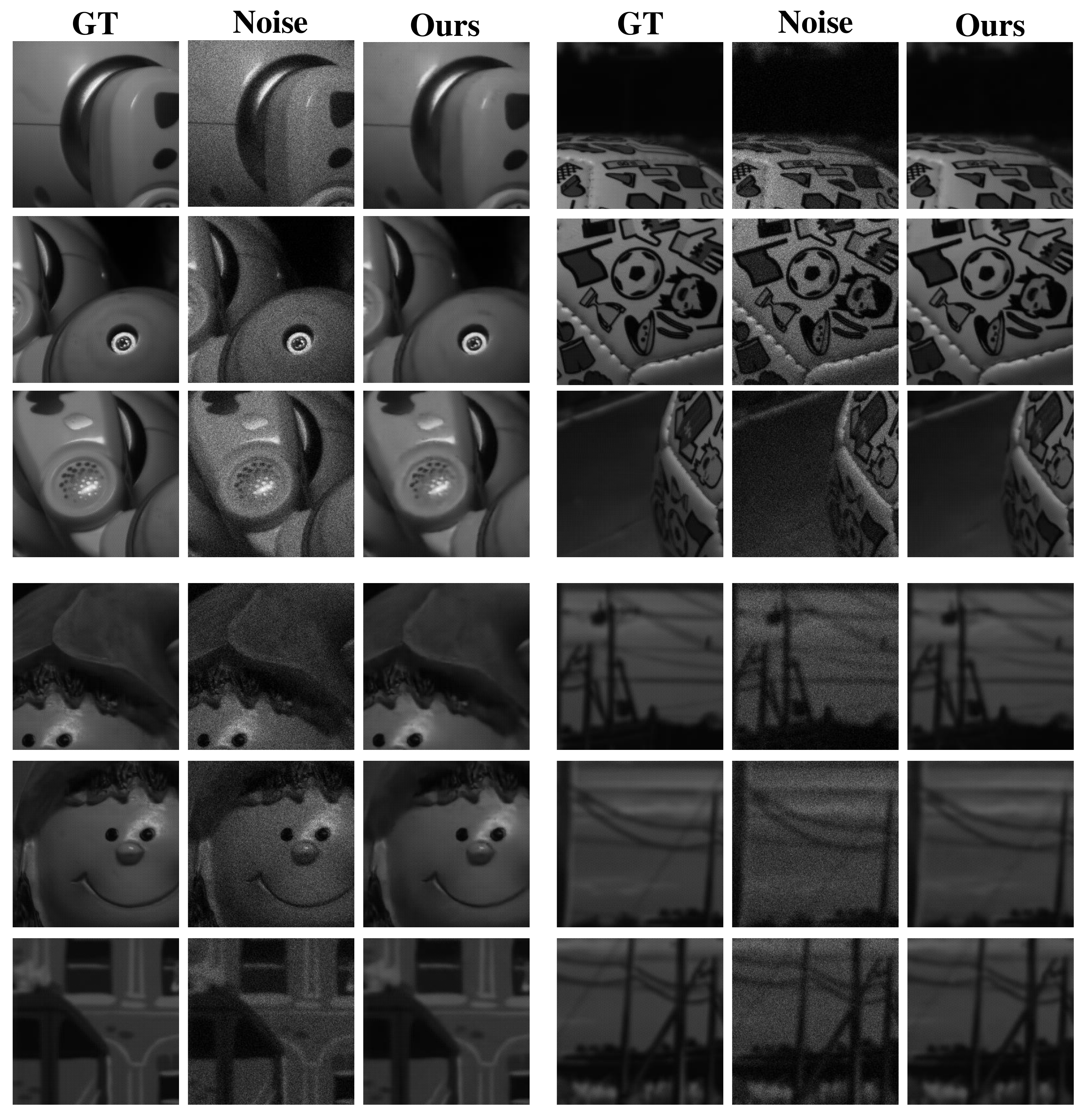}}
\end{center}
\vspace{-2.5mm}
\caption{Denoising visualized results under ISO 25600 settings on Raw videos dataset. 
}
\label{fig:9}
\end{figure}  
\noindent

\begin{figure*}[ht]
\begin{center}
\setlength{\fboxrule}{0pt}
\setlength{\fboxsep}{0cm}
\fbox{\rule{0pt}{0in} \rule{.0\linewidth}{0pt}
\hspace{-3.5mm}
     \includegraphics[width=1.0\linewidth]{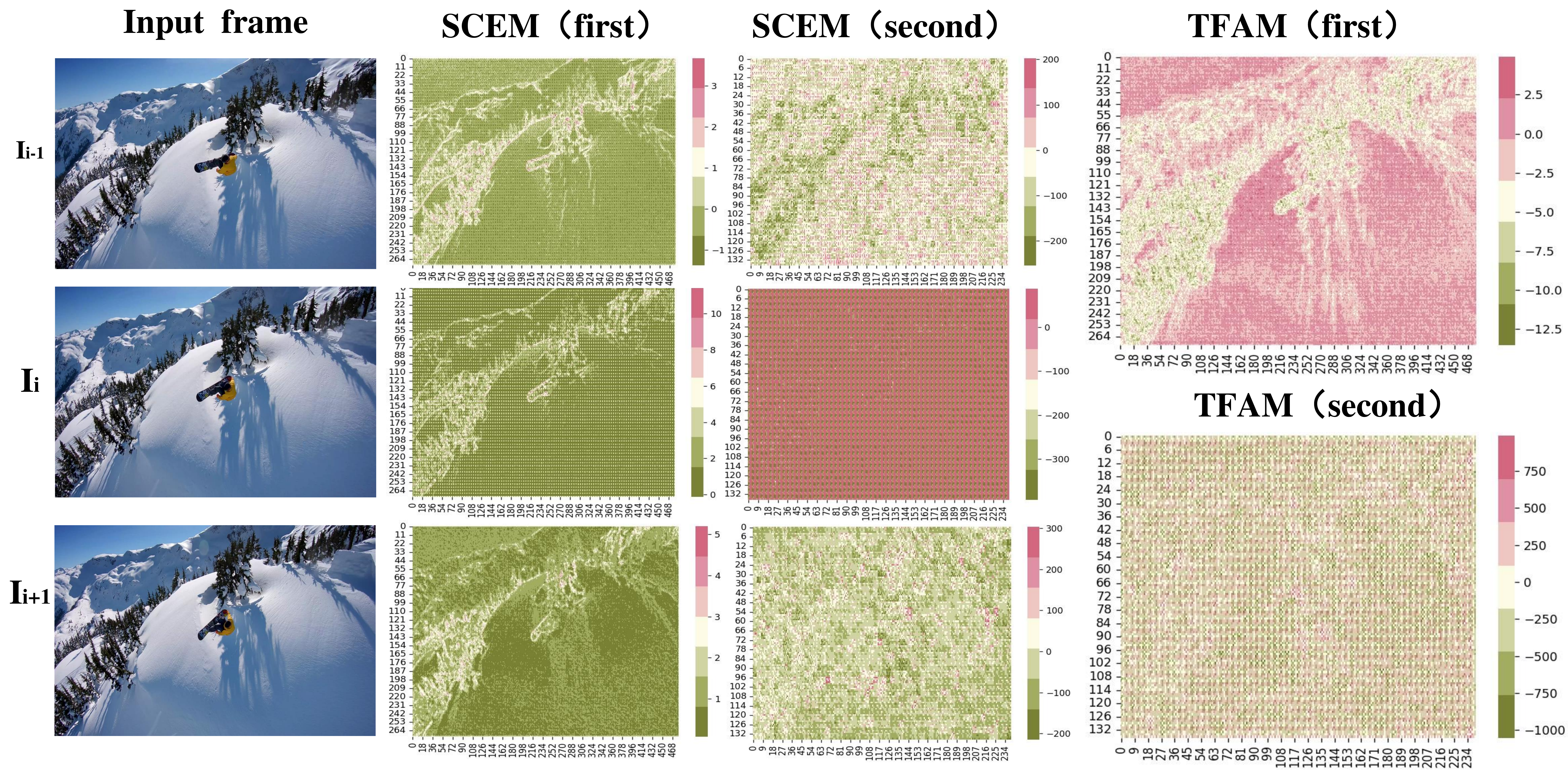}}
\end{center}
\vspace{-3.5mm}
\caption{Visualization outputs of SCEM and TFAM. We show the features of the consecutive input frames from Set8 dataset after twice pass through SCEM and TFAM during the coarse-level stages.}
\label{fig:10}
\end{figure*}  
\begin{figure}[htbp]
\begin{center}
\hspace{-5mm}
\setlength{\fboxrule}{0pt}
\setlength{\fboxsep}{0cm}
\fbox{\rule{0pt}{0in} \rule{.0\linewidth}{0pt}
\hspace{-2.5mm}
     \includegraphics[width=1.0\linewidth]{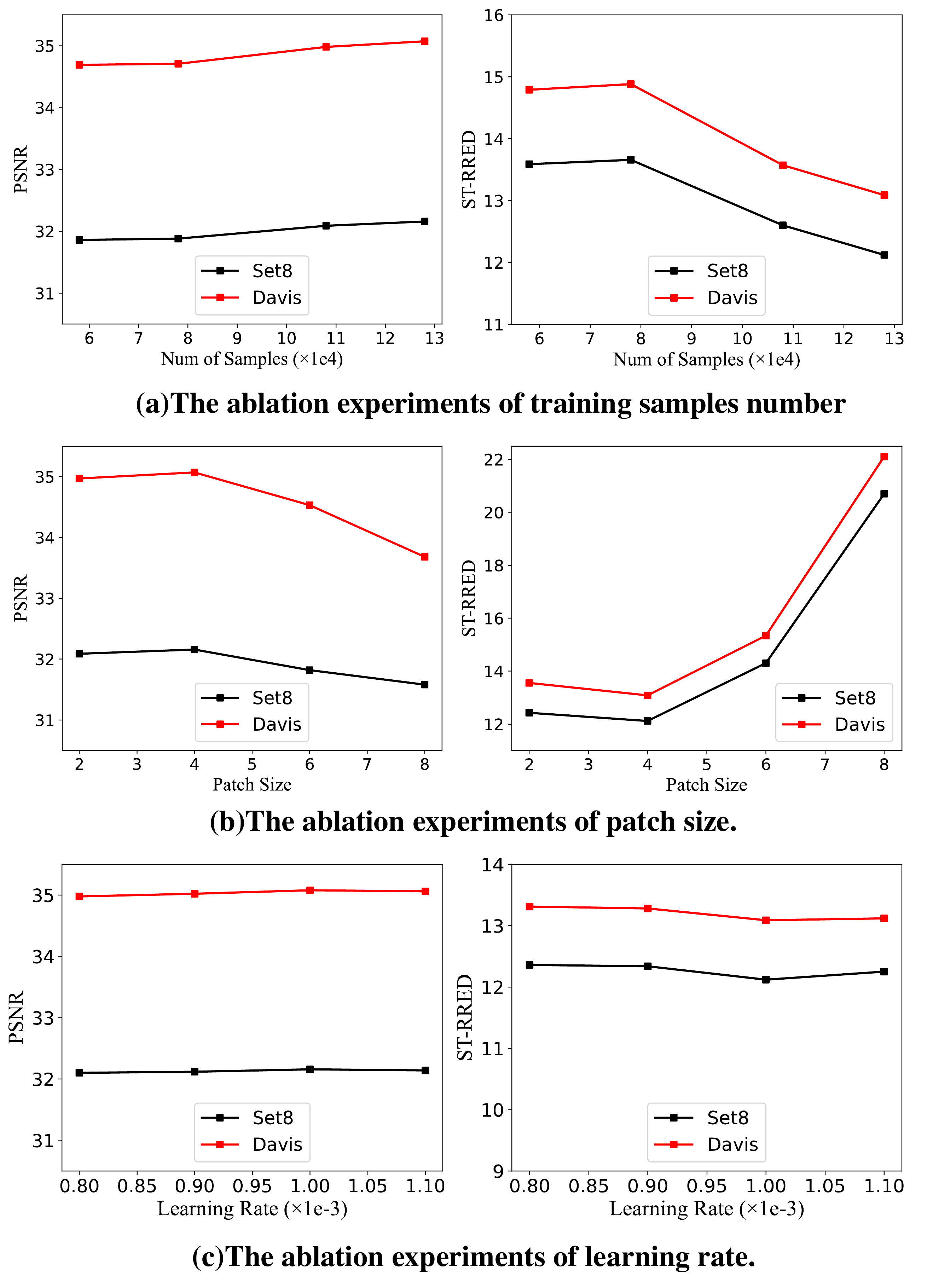}}
\end{center}
\vspace{-3.5mm}
\caption{Ablation study on the training samples number, patch size and learning rate. }
\label{fig:11}
\end{figure}


\textbf{Visualization results on Vimeo90k.}~ 
Figure ~\ref{fig:7}
shows the visualization results by comparing our proposed method with DnCNN~\cite{Zhang2017BeyondAG}, V-BM4D~\cite{Maggioni2012VideoDD}, TOFlow~\cite{xue2019video}, FastDVDnet~\cite{Tassano2020FastDVDnetTR} on Vimeo90k dataset with the Gaussian noise under standard deviation $\sigma=50$.

According to the visualization results, our proposed model can recover clearer details for nail caps and hand details than other state-of-the-art methods. The reasons stem from: firstly, we use a coarse-to-fine strategy to progressively implement coarse-grained to fine-grained denoising; secondly, our method inherits the advantages of both CNN and transformer in extracting local features and modeling long-range information compared to other methods; lastly, we preserve the important detailed information of the video during denoising.

\noindent
\textbf{Results on real noise.} 
To verify the effectiveness of our method with real noise, we further evaluate the proposed DSCT method on Raw Videos dataset. During the training and evaluation, we adopt the same training data and testing data followed~\cite{Sheth_2021_ICCV}, and optimize our model under each specific ISO setting. We compare our method to previous state-of-the-art methods including DIDN~\cite{9025411}, RViDeNet\cite{Yue_2020_CVPR} and UDVD~\cite{Sheth_2021_ICCV}. Among them, DIDN is a specific real-image denoising method. UDVD can directly optimize the denoising model on the Raw videos dataset. RViDeNet is pre-trained on synthetic data and then fine-tuned on the Raw video dataset. 
We present the results in Table~\ref{tab:3}, and we can see that our proposed method obtains better performance on Raw Videos dataset than other methods under different ISO level noisy types.
We can observe that our method outperforms the previous methods, such as DIDN~\cite{9025411}, RViDeNet~\cite{Yue_2020_CVPR} and UDVD~\cite{Sheth_2021_ICCV}, indicating our method can capture more effective global contextual dependencies to greatly beneficial the denoising results.

\begin{table}[ht]
\renewcommand{\arraystretch}{1.2}

\setlength{\tabcolsep}{2.9mm}{
\centering
\caption{PSNR comparison with our proposed method and other state-of-the-art denoising methods on Raw videos dataset. Different ISO represents various levels of noise.}
\vspace{-1.5mm}
\label{tab:3}
\begin{tabular}{|c|c|c|c|c|c|}
\hline
\diagbox[width=9em]{Method}{ISO}      & 1600  & 3200  & 6400  & 12800 & 25600 \\ \hline
DIDN~\cite{9025411}     & 47.00 & 45.02 & 43.08 & 40.58 & 40.56 \\ 
RViDeNet~\cite{Yue_2020_CVPR} & 47.74 & 45.91 & 43.85 & 41.20 & 41.27 \\ 
UDVD~\cite{Sheth_2021_ICCV}    & 48.04 & 46.24 & 44.70 & 42.19 & 42.11 \\ \hline 
\textbf{Ours}     & \textbf{48.97} & \textbf{47.11} & \textbf{45.61} & \textbf{43.04} & \textbf{43.01}  \\ \hline 
\end{tabular}
}
\end{table}

\noindent
\textbf{Visualization results on Raw videos.}~
In order to intuitively observe the effectiveness of our method, we further show more visualization results on the Raw videos dataset under ISO level setting 25,600 in Figure~\ref{fig:9}. The results show that our proposed method eliminates the noise yet retains the essential visual content, achieving better denoising results. 


\subsection{Ablation Study}
In this section, we conduct ablation studies to examine the impact of different components in our approach on Davis and Set8 datasets with Gaussian noise standard $\sigma$=30.

\noindent
\textbf{Impact of different stage.} 
In our proposed method, we propose a dual-stage method, i.e., the first stage is the \emph{coarse-level stage}, and the second stage is the \emph{fine-level stage}. We explore the impact of the \emph{coarse-level stage} and the \emph{fine-level stage}. One only has a coarse stage and the other only has a fine stage. The results are shown in Table~\ref{tab:4}. As listed in the table, the results of only the \emph{coarse-level stage} or the \emph{fine-level stage} are not satisfied, while the results of both stages achieve the best performance demonstrating the coarse-to-fine strategy is rewarding for the video denoising.
We explain the performance as the following reasons: the \emph{coarse-level stage} can extract dynamic features from spatio-temporal dimensions of neighbor frames and make full use of the temporal information, but it ignores detailed information in the process of denoising. The \emph{fine-level stage} can extract static features from the coarse-grained denoising, but if it is taken out separately, it only processes a single video frame and ignores the temporal relationship between video frames.
So, the two stages are complementary to each other in our task.

\noindent
\textbf{Effect of the Spatial-Channel Transformer.} 
We explore the effectiveness of the proposed Spatial-Channel Transformer in different modules. The ablation experimental results are shown in Table~\ref{tab:5}, in which ST denotes the swin transformer and SCEM denotes the proposed Spatial-Channel Encoding Module. We adopt the CNN architecture as the~\emph{base} model by only removing all transformer parts and the Temporal Features Aggregation Module~(TFAM) part. Following the base model, we progressively add different parts: firstly, we add a normal Swin transformer layer~\cite{9710580} as~\emph{base+ST} model, and then we replace it with our proposed Spatial-Channel Encoding Module denoted as~\emph{base+SCEM} model. In addition, we add TFAM to each of the above two models to obtain~\emph{base+ST+TFAM} model and our DSCT model. As shown in the table, after incorporating the transformer, the PSNR performance can be improved from 33.69 to 34.32 on Davis dataset, which veriﬁes the effectiveness of the modeling long-range dependency. When SCEM is incorporated, the PSNR performance can be improved from 31.66 to 31.94 on the Set8 dataset, showing the effectiveness of capturing long-range information from both spatial and channel.

\begin{table}[htbp]
\renewcommand{\arraystretch}{1.2}
 
\setlength{\tabcolsep}{3.2mm}{
\centering

\caption{PSNR/ST-RRED comparison among Coarse-level stage and Fine-level stage.}
\vspace{-2.5mm}
\label{tab:4}
\begin{tabular}{|c|c|c|c|}

\hline
Dataset  & Coarse-level stage & Fine-level stage & Ours \\ \hline
Davis   &    34.91/13.97                 &   33.79/23.34        &    \textbf{35.08/13.09}\\ 
Set8    &    32.07/13.03               &      31.35/21.68                       & \textbf{32.16/12.12}                \\ \hline

\end{tabular}
}
\end{table}

\begin{table}[htbp]
\renewcommand{\arraystretch}{1.2}
\setlength{\tabcolsep}{1.8mm}{
\centering
\caption{Ablation study on different modules. Test on Davis and Set8 datasets with a noise standard of 30. ST denote SwinTransformer, SCEM denote Spatial-Channel Encoding Module, TFAM denote Temporal Features Aggregation Module.}
\vspace{-1.5mm}
\label{tab:5}
\begin{tabular}{|c|c|c|c|cc|}

\hline
\multicolumn{1}{|c|}{\multirow{2}{*}{Method}} & \multicolumn{1}{c|}{\multirow{2}{*}{ST}} & \multicolumn{1}{c|}{\multirow{2}{*}{SCEM}} & \multicolumn{1}{c|}{\multirow{2}{*}{TFAM}} & \multicolumn{2}{c|}{PSNR/ST-RRED}  \\ \cline{5-6} 
\multicolumn{1}{|c|}{}                        & \multicolumn{1}{c|}{}                    & \multicolumn{1}{c|}{}                     & \multicolumn{1}{c|}{}                     & \multicolumn{1}{c|}{Davis} &  \multicolumn{1}{c|}{Set8} \\ \hline
base                                         &                                          &                                           &                                           & \multicolumn{1}{c|}{33.69/15.60 } &  31.24/14.36     \\ 
base+ST                                      &  \; $\checkmark$                         &                                           &                                           & \multicolumn{1}{c|}{34.32/16.46}      & 31.66/15.43      \\ 
base+ST+TFAM                                      &  \; $\checkmark$                         &                                           &  \; $\checkmark$                                         & \multicolumn{1}{c|}{34.80/14.11}      & 31.99/13.15      \\ 
base+SCEM                                   &                                          &     \; $\checkmark$                       &                                           & \multicolumn{1}{c|}{34.75/14.21}      & 31.94/12.99      \\ \hline
\textbf{Ours}                                         &                                          &     \; $\checkmark$                       &    \; $\checkmark$                        & \multicolumn{1}{c|}{\textbf{35.08/13.09} }      & \textbf{32.16/12.12}     \\ \hline

\end{tabular}
}
\end{table}


\begin{table}[htbp]
\renewcommand{\arraystretch}{1.2}
\setlength{\tabcolsep}{2.5mm}{
\centering
\caption{PSNR/ST-RRED comparison among different skip-connection operation, where w/o F-Skip denotes the model without F-skip operation,w/o CF-Skip denotes the model without CF-skip operation, and w/o TFAM denotes the model without TFAM.}
\vspace{-1.5mm}
\label{tab:6}
\begin{tabular}{|c|c|c|c|c|}
\hline
Dataset  & w/o F-SKip   & w/o CF-SKip  & w/o TFAM & Ours\\ \hline
Davis      &   35.01/13.57         &   34.15/22.10   &   34.16/14.51  & \textbf{35.08/13.09} \\ 
Set8    &  \textbf{32.16}/12.20     &  31.58/20.82    &        31.94/12.99     &\textbf{32.16/12.12} \\ \hline
\end{tabular}
}
\end{table}

\noindent
\textbf{Impact of the Multi-scale Residual Structure.}
We conduct ablation experiments to verify the effectiveness of each skip-connection operation in our model, by removing other skip-connection operations separately. The experimental results are shown in Table~\ref{tab:6} under the Gaussian noise with standard $\sigma$=30. We denote the skip-connection operation in \emph{fine-level stage} as F-SKip, the skip-connection operation between \emph{coarse-level stage} and \emph{fine-level stage} as CF-SKip, and the skip-connection operation in \emph{coarse-level stage} as TFAM. From Table~\ref{tab:6}, we can observe that the performance degrades when each skip-connection operation is removed, so our proposed three skip-connection operations are an integral part to bring significant performance improvement.

In addition, we have validated different methods for aggregating multi-frame temporal information, such as fusing features using the mean method, convolution operation, and our proposed TFAM. The experimental results are shown in Table \ref{tab:7}, and we can see that our proposed TFAM beats other methods across both metrics, \emph{i.e.}, PSNR and ST-RRED. It demonstrates that our proposed TFAM can fully integrate the temporal features of different frames compared with the simple Mean or Conv operations.

\begin{table}[htbp]
\renewcommand{\arraystretch}{1.2}
\setlength{\tabcolsep}{4.5mm}{
\centering
\caption{PSNR/ST-RRED comparison among different aggregation methods, \emph{i.e.}, Mean, Conv operation and Temporal Features Aggregation module.}
\vspace{-1.5mm}
\hspace{-1.5mm}
\label{tab:7}
\begin{tabular}{|c|c|c|c|}
\hline
Dataset  & Mean  & Conv  & \textbf{TFAM (Ours)} \\ \hline
Davis    & 34.14/22.02    & 34.52/15.93      & \textbf{35.08/13.09}                     \\ 
Set8      &  31.55/20.86 & 31.75/15.12       & \textbf{32.16/12.12}                  \\ \hline
\end{tabular}
}
\end{table}

\noindent
\textbf{Visualization results of different modules.} To further verify the validity of each module, we visualized the features output of the consecutive input frames after the first pass during the \emph{coarse-level stage} through the SCEM and the TFAM, as well as the output results of the second pass through SCEM and TFAM during the \emph{coarse-level stage}. 
These results are shown in Figure~\ref{fig:10}, and we show the results on three consecutive sequences from Set8 dataset. According to the visualization results, we can observe that the output features through the first pass SCEM and TFAM contain the low-level spatial information of the video frames, and the output features after the second pass SCEM and TFAM contain the high-level semantic context of the video frames. Furthermore, we observe that the SCEM focuses on extracting more detailed spatial information, while TFAM mainly focuses on capturing the background information. 

\noindent
\textbf{The computation cost. } 
We report the computation cost (i.e., FLOPS) on Davis dataset with a noise standard of 30 in the table \ref{tab:8}. Our FLOPS is much lower than UDVD\cite{Sheth_2021_ICCV} and comparable to FastDVDnet~\cite{Tassano2020FastDVDnetTR}, but our method can achieve better PSNR performance, demonstrating our method obtains good trade-off between inference efficiency and performance.
 
\begin{table}[ht]
\renewcommand{\arraystretch}{1.2}

\setlength{\tabcolsep}{8.0mm}{
\centering
\caption{FLOPS and PSNR comparison with our proposed method and other state-of-the-art denoising methods on Davis dataset Gaussian noise level $\sigma$ = 30.}
\vspace{-1.5mm}
\label{tab:8}
\begin{tabular}{|c|c|c|}
\hline
Method      & FLOPS  & PSNR  \\ \hline
FastDVDnet~\cite{Tassano2020FastDVDnetTR} & \text{ 5.88G}   &34.04 \\
UDVD\cite{Sheth_2021_ICCV}  & 73.59G  &33.92 \\ \hline
Our   & 8.75G   & \text{35.08} \\ \hline
\end{tabular}
}
\end{table}

\noindent
\textbf{The number of training samples and patch size.} 
We conduct ablation experiments w.r.t the number of training samples and the patch size of the transformer, and Figure \ref{fig:11} shows the corresponding experimental results. Figure\ref{fig:11}~(a) shows the experimental results of various numbers of training samples. We increase the number of training samples sequentially from 58,000 to 128,000 in a span of 20,000. As expected, the performance of our model increases along with the number of training samples. In addition, from Figure\ref{fig:11}~(b), we can see that the best performance is achieved when we set the patch size as 4$\times$4, demonstrating the too large patch size is suitable for visual recognition instead of video denoising, \emph{e.g.}, VIT~\cite{kolesnikov2021image} set patch size to 16$\times$16 for classification. While the too small patch size is not suitable for our task, because the interaction between pixel points ignores the corresponding texture information. Therefore, the patch size of 4$\times$4 ensures the best results are achieved by focusing on the location information of the noise and texture details.

\noindent
\textbf{Impact of the learning rate.}~Different learning rates have various effects on the performance of our proposed model. Here we evaluate our model under different learning rates~(0.0008, 0.0009, 0.0010 and 0.0011) on Davis and Set8 datasets with Gaussian noise standard $\sigma=30$. The experimental results are shown in Figure \ref{fig:11}~(c). We can observe that our proposed method obtains the best performance when the learning rate is set as 0.001. Moreover, we also find that our method exhibit stable performance under different learning rates, demonstrating the robustness of our proposed model.

\section{Conclusion}
In this paper, we propose a novel Dual-stage Spatial-Channel Transformer, which uses the progressive coarse-to-fine strategy for video denoising. We design a Spatial-Channel Transformer to simultaneously capture spatio-temporal features, and a Spatial-Channel Encoding Module to build long-range spatial-channel information. In addition, a Multi-Scale Residual Structure is introduced to preserve the video detail information and enhance the representation on each level, which contains a Temporal Features Aggregation Module to summarize the dynamic representation. Extensive experiments on four public benchmarks demonstrate the effectiveness and superiority of the proposed method.

{\bibliography{main}
\bibliographystyle{IEEEtran}}

\end{document}